\documentclass[sn-basic]{sn-jnl}
\usepackage{threeparttable}
\normalbaroutside 
\usepackage{changes}
\usepackage{parskip}
\usepackage{mathtools}
\usepackage{placeins}
\usepackage{booktabs}
\usepackage{bold-extra}

\newcommand{\dparagraph}[1]{\textbf{#1} \mbox{}}
\jyear{2023}
\theoremstyle{thmstyleone}

\theoremstyle{thmstyletwo}

\theoremstyle{thmstylethree}

\begin{document}
\title[ ]{Effect of hyperparameters on variable selection in random forests}
\author*[1,2]{\fnm{Cesaire J.} \sur{K. Fouodo}}\email{cesaire.kuetefouodo@uni-luebeck.de}
\author[1]{\fnm{Lea L.} \sur{Kronziel}}\email{l.kronziel@uni-luebeck.de}
\author[1,2]{\fnm{Inke R.} \sur{K\"onig}}\email{inke.koenig@uni-luebeck.de}
\author*[1,2]{\fnm{Silke} \sur{Szymczak}}\email{silke.szymczak@uni-luebeck.de}
\affil*[1]{\orgdiv{Institut für Medizinische Biometrie und Statistik}, \orgname{Universit\"at zu L\"ubeck and Universit\"atsklinikum Schleswig-Holstein, Campus L\"ubeck}, \orgaddress{\street{Ratzeburger Allee}, \city{L\"ubeck}, \postcode{23562}, \state{Schleswig-Holstein}, \country{Germany}}}
\affil[2]{\orgdiv{DZHK (German Center for Cardiovascular Research)}, \orgname{partner site L\"ubeck}}

\abstract{Random forests (RFs) are well suited for prediction modeling and variable selection in high-dimensional omics studies. The effect of hyperparameters of the RF algorithm on prediction performance and variable importance estimation have previously been investigated. However, how hyperparameters impact RF-based variable selection remains unclear. We evaluate the effects on the Vita and the Boruta variable selection procedures based on two simulation studies utilizing theoretical distributions and empirical gene expression data. We assess the ability of the procedures to select important variables (sensitivity) while controlling the false discovery rate (FDR).

Our results show that the proportion of splitting candidate variables (\texttt{mtry.prop}) and the sample fraction (\texttt{sample.fraction}) for the training dataset influence the selection procedures more than the drawing strategy of the training datasets and the minimal terminal node size. A suitable setting of the RF hyperparameters depends on the correlation structure in the data. For weakly correlated predictor variables, the default value of \texttt{mtry} is optimal, but smaller values of \texttt{sample.fraction} result in larger sensitivity. In contrast, the difference in sensitivity of the optimal compared to the default value of \texttt{sample.fraction} is negligible for strongly correlated predictor variables, whereas smaller values than the default are better in the other settings.

In conclusion, the default values of the hyperparameters will not always be suitable for identifying important variables. Thus, adequate values differ depending on whether the aim of the study is optimizing prediction performance or variable selection.}


\maketitle

\section{Introduction}\label{sec1}

Random forests (RFs) have received much attention in recent years. They have been first introduced by \cite{breimanrandom2001} for classification and regression problems before being later extended by \cite{ishwaranRSF2008} for survival analyses. They are a powerful predictive method, do not make assumptions on the distribution of predictor variables, and can capture variable importance. In life science and medicine, RFs are commonly used to analyze high dimensional omics data, either for prediction or explanation purposes, for example, to understand the biological pathways leading to a disease of interest. 

To estimate variable importance with RFs for classification problems, the impurity importance measure, often called the Gini importance measure, is commonly used \citep{breimanrandom2001}. One of its particular advantages is that it can be computed quickly compared to other RF-based variable importance measures like the permutation variable importance or the \textit{holdout} variable importance measure \citep{janitza2016}. However, it has been shown that the Gini importance measure can be biased since it favors variables with a larger number of split points \citep{stroblbias2007,boulesteixrandom2012}. To overcome this limitation, the corrected Gini importance has been implemented by \cite{nembrinirevival2018}. In general, variable importance values can be used to rank predictor variables. However, to distinguish effect from noise predictor variables, RF variable selection methods have emerged in recent years. In a systematic comparison study, \cite{degenhardtevaluation2019} have compared different approaches and recommended the Vita \citep{janitza2016} and the Boruta \citep{kursafeature2010} method. 

Similarly to other machine learning algorithms, RFs have several hyperparameters that can be tuned for optimal performance on a specific dataset. For example, to achieve a good prediction accuracy, some authors argued that the number of trees (\texttt{num.trees}) should be increased \citep{diazuriartegene2006,refId0,oshiro2012}. In contrast, decreasing the proportion of predictor variable candidates (\texttt{mtry}) used for node splitting and the proportion of the training dataset used for growing each tree (\texttt{sample.fraction}) reduces the correlation between the trees, resulting in more diverse prediction results across trees, and a more stable prediction after aggregation over the forest \citep{probsthyperparameters2019}. According to \cite{martinez2010}, if the hyperparameter \texttt{sample.fraction} is optimally set, the sampling strategy \texttt{replace}, indicating whether the training dataset is sampled with replacement, will not have a large influence on the prediction performance. \cite{stroblbias2007} and \cite{martinez2010} recommend to draw the training dataset with replacement. However, drawing with replacement can bias the prediction if the dataset simultaneously contains quantitative and qualitative predictor variables. The minimal terminal node size (\texttt{min.node.size}) is crucial for the depth of the trees. According to \cite{segal2004,stroblbias2007}, the more noise variables are expected in the dataset, the more accurate the RF will be with a higher  \texttt{min.node.size} value.

In addition to optimizing prediction accuracy, hyperparameters also influence variable importance estimation. The optimal setting depends on the data structure and the study's goal. If the dataset includes correlated predictor variables, and the goal is to identify all variables in relationship with the response variable, a low number of splitting candidate variables is preferred because the variables with moderate or weak effects will receive a higher chance to be selected \citep{grompingvariable_2009, probsthyperparameters2019}. However, if a minimal set of required variables with a high prediction accuracy is the focus, increasing the number of splitting candidate variables will favor predictor variables with strong effects and, therefore, underestimates the importance of those with moderate or weak effects on the response variable \citep{genuervariable2010}. In addition, \cite{probsthyperparameters2019} recommend a small sample fraction to train the decision trees. \cite{grompingvariable_2009,genuervariable2010} argued that a small minimal terminal node size is advantageous for those predictor variables with moderate or weak effects on the response variable. Understandably, keeping the minimal node size small should increase splitting possibilities and, therefore, the chances of using variables with small effects. According to \cite{probsthyperparameters2019}, the effect of the drawing strategy on the RF variable importance measures is similar to those on predictive performances, i. e. drawing with replacement is preferred. Finally, growing many trees is generally recommended for better stability \citep{lunettascreening2004,genuervariable2010}.

The impact of RF hyperparameters on the variable selection methods has received less attention. In their comparison study, apart from the number of trees, \citep{degenhardtevaluation2019} used the default RF hyperparameters setting of the \texttt{R} package \texttt{ranger} \citep{ranger}. In addition to the number of trees, \cite{janitza2016} varied only the number of split candidate variables while setting the other hyperparameters as default. Finally, \cite{kursafeature2010} did not investigate the impact of RF hyperparameters on the Boruta method. Since it has been established that hyperparameters can influence RF variable importance measures, it is important to understand their possible impact on the resulting variable selection methods.

We thus investigate the effect of the RF hyperparameters on the Vita and Boruta variable selection procedures and give some recommendations for optimal hyperparameter settings. We perform two simulation studies of a classification problem with a dichotomous response variable and high dimensional correlated predictor variables. In the first study, our focus is on block-wise correlation structures, while in the second study, more complex and more realistic correlation patterns are estimated from experimental gene expression data.

\section{Materials and methods}\label{sec2}

\subsection{Random forests}
\label{subsec:rf}
\noindent
Random forests (RFs) are ensemble methods of decision trees \citep{breimanrandom2001}. A sample fraction of the original data, drawn with or without replacement, defined as a training dataset, is used to generate each tree. The remaining observations, considered as the testing dataset, can be used to estimate the prediction error of the resulting tree. Each tree is generated by the recursive splitting of the training dataset into two child nodes until a stopping criterion, such as tree depth, minimal node size, or node impurity, is achieved. A subset of variable candidates is randomly drawn at each splitting step, and the one maximizing the impurity decrease based on the Gini index is used to split the current node. 

\subsection{Variable importance and  selection methods}
\label{subsec:vimp}
\noindent
One of the main advantages of RFs is that they can be used to estimate the variable importance of the predictor variables. For instance, the Gini importance measure determines each predictor variable's mean decrease in impurity when the corresponding variable is used for splitting. Because this measure is biased for classification problems, the corrected Gini importance is considered for this study, and we use the implementation proposed by \cite{nembrinirevival2018}, available in the \texttt{R} package \texttt{ranger} \citep{ranger}. One property of the corrected Gini importance measure is that the resulting values can be negative, equal to zero, or positive. Therefore, they are interpretable for variable selection. Based on the estimated variable importance, RF variable selection methods allow distinguishing noise variables (not being in relationship with the response variable) from effect variables (being in relationship with the response variable). Because \cite{degenhardtevaluation2019} has recommended Vita \citep{janitza2016} and Boruta \citep{kursafeature2010} as the most powerful methods, we use these two methods in this study. The two methods have been wrapped in the \texttt{R} package \texttt{Pomona} available on \url{https://github.com/imbs-hl/Pomona}. 

\subsubsection{Vita}
\noindent
The Vita variable selection method is a testing procedure that estimates a $p$-value for each predictor variable based on the estimated variable importance and an estimated null distribution \citep{janitza2016}. The null distribution is created based on predictor variables with no relationship to the response variable. For this, the estimated non-positive variable importance values are mirrored around zero to get a distribution based on which the $p$-value for each predictor variable is estimated. Given a significance threshold, a predictor variable is claimed to be important for prediction if the resulting $p$-value is smaller or equal to the significance threshold; otherwise, it is unimportant and considered a noise variable. Using the standard significance level of $5$\% will result in a high false discovery rate (FDR). Thus,  \cite{degenhardtevaluation2019} used a very stringent criterion, i.e. only predictor variables with an empirical $p$-value of zero were selected. Here we use a less conservative strategy by applying the Benjamini-Hochberg (BH) procedure \citep{benjaminicontrolling1995} to adjust for the multiple tests.

\subsubsection{Boruta}
\noindent
The Boruta method does not estimate $p$-values but categorizes the predictor variables into the three groups \texttt{confirmed}, \texttt{tentative} and \texttt{rejected} using an iterative algorithm \citep{kursafeature2010}. Initially, the algorithm sets all predictor variables to \texttt{tentative}. At each iteration, the original dataset is extended by shadow variables created by permuting the values of the original predictor variables. Thus, shadow variables are not intended to be related to the response variable. Boruta determines the importance of original and shadow variables at each iteration. For each predictor variable it is counted in how many iterations the estimated importance is larger than the maximum value estimated among the shadow variables. Then a binomial test with a given significance level is applied to test if the predictor variable has significantly larger or smaller counts than expected by chance and is then denoted as \texttt{confirmed} or \texttt{rejected}. The latter variables are removed from further iterations. The algorithm stops when all predictor variables are \texttt{confirmed} or \texttt{rejected}, or the maximum number of iterations is reached. At each iteration, we use the Bonferroni correction \citep{blandmultiple1995} to control the family-wise error rate.

\subsection{RF hyperparameters}
\label{subsec:hyperparam}
\noindent

The RF hyperparameters considered in this study are summarized in Table \ref{table:hyperparam}, together with their default values as implemented in the \texttt{R} package \texttt{ranger}. The default value of \texttt{mtry} is set to $1/3$ for regression \citep{randomforest}. Note that \texttt{mtry} and \texttt{min.node.size} are given as absolute numbers. However, in the following, these hyperparameters will be defined as proportions of the number of predictor variables and of the sample size.

\begin{table}[htbp]
	\centering
	\footnotesize
  \begin{threeparttable}
  	\caption{Overview of RF hyperparameters investigated in this study}
\begin{tabular}{llp{3cm}p{4cm}}
	\toprule
	\textbf{Hyperparameter$^a$} 	& \textbf{Type} 	& \textbf{Description} & \textbf{Default value$^b$} \\
	\midrule
		\texttt{num.trees} 			& integer 			& Number of trees  & $500$ \\
		\texttt{mtry} 				& numeric 			& Number of candidate predictor variables for splitting at each node & $\sqrt{p}$, with $p=$ number of predictor variables \\
	\texttt{replace} 			& logical 			& Sample with replacement  & \texttt{TRUE} \\
	\texttt{sample.fraction} 	& numeric 			& Number of samples to be drawn for each tree & $1$ if $\texttt{replace} = \texttt{TRUE}$, $0.632$ else \\
	\texttt{min.node.size} 		& integer 			& Minimal terminal node size & $1$ for classification, $5$ for regression, with $n=$ sample size \\
	\bottomrule
\end{tabular}
    \begin{tablenotes}
      \footnotesize
      \item $^{a, b}$ as in \texttt{R} package \texttt{ranger}
    \end{tablenotes}
  \end{threeparttable}
  \label{table:hyperparam}
\end{table}

\subsection{Simulation studies}
\label{subsec:simul}
\noindent
We conduct two simulation studies. The first one is based on theoretical distributions where the correlation of the effect variables is defined in a simple block structure. The advantage of this setting is that results can be interpreted with regard to the correlation's strength. The second study uses empirical gene expression data to mimic realistic complex correlation patterns. Both studies' descriptions follow the ADEMP scheme \citep{morris2019}.

\subsubsection{Simulation study 1: Simple correlation structure}
\label{sim:study1}

\hypertarget{study1:aim}{
\label{chap:hyperparameter:simulationsstudie1:aim}}
\dparagraph{Aim}
This simulation study aims to assess how well the two variable selection methods, Vita and Boruta, can distinguish between noise and effect predictor variables under different hyperparameter settings in datasets with simple correlation patterns.

\hypertarget{study1:data}{
\dparagraph{Data generating mechanism}}
Predictor variables are generated following the approach used in simulation study 1 of \citet{degenhardtevaluation2019}. Six independent base variables $x_q$ are simulated from the uniform distribution $\mathcal{U}(0, 1)$. Subsequently, each base variable is used to generate blocks of correlated variables. The blocks are simulated so that the correlation decreases with an increasing number of predictor variables. The $j$th correlated variable derived from the base variable $x_q$, denoted as $v^{(j)}_q$, is defined as:

\begin{align*}
  v^{(j)}_q = x_q + \left(0.01 + \frac{0.5(j - 1)}{k - 1}\right)\times \eta
\end{align*}

where $j = 1, \dots, k$ and $\eta$ is a realization from $\mathcal{N}(0, 0.3)$, $0.3$ being the standard deviation. The form of $v^{(j)}_q$ indicates a relationship between $v^{(j)}_q$ and $x_q$, and increasing values of $j$ lead to lower correlations. For fixed values of $j$, the correlation is stronger for smaller values of $k$. To increase the number of predictor variables $p$ to $5000$, additional independent predictor variables are added, drawn from $\mathcal{U}(0, 1)$. 

In contrast to \citet{degenhardtevaluation2019}, we simulate a dichotomous response variable instead of a quantitative one. The response variable is simulated using the first three base variables, not their corresponding correlated variables. For an observation $\mathbf{x}^{\textsc{T}} = \left(v^{1}_1, v^{1}_2, \dots, v^{1}_6, v^{2}_1, v^{2}_2, \dots, v^{2}_6, \dots, v^{k}_1, v^{k}_2, \dots, v^{k}_6, x_{6k + 1}, \dots, x_p\right)$, the response variable is defined based on a logistic regression model without intercept, i.e. the success probability is defined as:

\begin{align*}
	\text{Pr}(Y = 1 \rvert \mathbf{x}) = \frac{\exp({x_1\beta_1 + x_2\beta_2 +  x_3\beta_3})}{1 + \exp(x_1\beta_1 + x_2\beta_2 + x_3\beta_3)},
\end{align*}

where $\beta_1, \beta_2,$ and $\beta_3$ represent the effects of the first three base variables on the response variable. Note that the correlated variables $v{(j)}_q$'s are included in the final dataset instead of the base variables $x_1, x_2$, and $x_3$. The effects for the noise variables are set to $0$. Effect sizes $\beta$ are randomly drawn from the set $\left\{-3, 3, -2, 2, -1, 1, 0\right\}$ for each replicate separately. We use two different block sizes ($k\in{10, 50}$) and generate $100$ replicates with  $n = 100$ observations. 

\hypertarget{study1:estimand}{
\dparagraph{Estimands}}
We record the important predictor variables for each variable selection method and hyperparameter setting.

\hypertarget{study1:method}{
\dparagraph{Methods}}
We compare different hyperparameter settings for the two variable selection methods, Vita and Boruta. For Vita, predictor variables with a BH corrected $p$-value smaller than $0.05$ are defined as important, whereas only variables in the category \texttt{confirmed} are used for Boruta. Both methods are applied using the \texttt{R} package \texttt{Pomona} version 1.0.2, which is based on implementations provided by the \texttt{R} packages \texttt{ranger} version 0.12.1 and \texttt{Boruta} version 7.0.0. All method-specific hyperparameters (for Vita, the significance threshold of $0.05$ and, for Boruta, the threshold of $0.01$) are set to their default values. 

The different RF hyperparameters are varied according to the values listed in Table \ref{chap:hyperparam:sim1:var1}. When varying one of the hyperparameters, the other ones are kept fixed to their default values. The number of trees (\texttt{num.trees}) is kept constant and set to a large value of $10,000$. The hyperparameter combination $\texttt{replace} = \texttt{FALSE}$ and $\texttt{sample.fraction} = 1$ is excluded since it would use all observations for training. Job parallelization is performed using the \texttt{batchtools} \texttt{R} package version 0.9.13 \citep{langbatchtools2017} on a 64-bit Linux platform with two 16-core Intel Xeon ES-2698 v3 CPUs, with three CPUs per job.

\begin{table}[htbp!]
	\centering
  \begin{threeparttable}
	\caption{Variations of hyperparameters in simulation study 1 with simple correlation structure. The default values of each hyperparameter are highlighted in bold.}
	\begin{tabular}{ll}
		\toprule
		\textbf{Hyperparameter} & \textbf{Variation} \\
		\toprule
		\texttt{mtry.prop}$^a$ & $\textbf{0.014}\ 0.1,\ 0.2,\ 0.33,\ 0.5$\\
		\texttt{replace} & \texttt{\textbf{TRUE}};\ \ \texttt{FALSE}\\
		\texttt{sample.fraction} & $0.2,\ \ 0.4,\ \ \mathbf{0.632},\ \ 0.8,\ \ 1.0$\\
		\texttt{min.node.size.prop}$^b$ & $0.01,\ 0.05,\ 0.1,\ 0.2,\ \ \mathbf{1/n}$\\
		\bottomrule
	\end{tabular}
	\label{chap:hyperparam:sim1:var1}
    \begin{tablenotes}
      \footnotesize 
      \item $^a$ given as a proportion of the number of predictor variables
      \item $^b$ given as a proportion of the total sample size
    \end{tablenotes}
\end{threeparttable}
\end{table}

\hypertarget{study1:performance}{
\dparagraph{Performance measures}}
To evaluate the performance of the two variable selection methods, the variables correlated to the first three base variables, i.e. those with an effect on the response variable, are defined as true predictor variables. Consequently, all other predictor variables are considered as noise variables. Three different performance measures are estimated. The first two are estimated based on each single replicate. The FDR is the proportion of falsely selected predictor variables among the variables identified by the variable selection method. The sensitivity is estimated by dividing the number of selected important variables by the total number of the simulated effect predictor variables ($= 30$ or $150$ for $k = 10$ or $50$). The third performance is stability which compares the similarity of lists of important variables across replicates. For each pair of replicates, stability is determined using Jaccard’s index, which is calculated as the division of the length of the intersection and the length of the union of the two lists of important variables \citep{hestable2010}. The index ranges from a minimum value of $0$ to a maximum of $1$, with higher values indicating better stability. The average index is calculated across all possible pairs or replicates.

\subsubsection{Simulation study 2: Empirical correlation structure}
\label{sim:study2}

\hypertarget{study2:aim}{
\dparagraph{Aim}}
This simulation study aims to evaluate the effect of RF hyperparameters on the variable selection with Vita and Boruta in datasets with realistic complex correlation patterns.

\hypertarget{study2:data}{
\dparagraph{Data generating mechanism}}
Similar to previous studies \citep{janitza2016,nembrinirevival2018}, the evaluation of the variable selection methods in this simulation study is based on an empirical gene expression dataset. This dataset is accessible from the \texttt{R} package \texttt{cancerdata} \citep{cancerdata} under the name \texttt{VEER1}. It contains gene expression data from the breast cancer microarray study of \cite{veer2002}, filtered by \cite{michiels2005}. We removed two additional genes because of missing values. The resulting dataset contains gene expression values for $4946$ genes in $78$ tumor samples. Each predictor variable is standardized to a standard deviation of $1$ to facilitate the comparison of the simulated effects.

The original response variable is ignored, and the following steps are performed for each of the $100$ replicates. First, $50\%$ of the patients are randomly selected as cases, and the remaining ones are set to be controls. Second, $200$ predictor variables are randomly selected to be effect variables, and each of the effect sizes out of the set of $\{-0.1;\ 0.1;\ -0.2;\ 0.2;\dots;-0.7;\ 0.7;\ -0.8;\ 0.8\}$ is assigned to 20 of the effect variables. Effect sizes $\rvert 0.1 \rvert$, $\rvert 0.2 \rvert$ are considered as weak, $\rvert 0.3 \rvert$, $\rvert 0.4 \rvert$, $\rvert 0.5 \rvert$ as moderate and $\rvert 0.6 \rvert$, $\rvert 0.7 \rvert$ and $\rvert 0.8 \rvert$ as strong. The effects are introduced by a mean shift, i.e., the corresponding effect size is added to cases and subtracted from controls. In order to be able to identify false positive results, correlation between true and noise variables as well as within noise variables is removed by permuting the values of each of the noise variables, separately.

\hypertarget{study2:estimand}{
\dparagraph{Estimands}}
The same estimand as in simulation study 1 is used.

\hypertarget{study2:method}{
\dparagraph{Methods}}
The same methods and hyperparameter settings as in simulation study 1 are used. However, in this simulation study, we set the total number of trees to $14,844$. Analogous to simulation study 1, only one of the hyperparameters is varied at a time while the others are set to their default values.

\hypertarget{study2:performance}{
\dparagraph{Performance measures}}
True effect and noise variables are defined as described in the paragraph data generating mechanism. Sensitivity and FDR are used as performance measures again. We excluded the stability calculation since the effect variables differ from one replicate to another.

\section{Results}
\label{sec:res}

For each simulation study, we show the results of the most influential hyperparameters \texttt{mtry.prop} and \texttt{sample.fraction} in the main text, while figures for the other hyperparameters are presented in the supplement. Each subfigure compares the values of one of the evaluation criteria for varying values of one of the hyperparameters while the other hyperparameters are set to their default values. 

\subsection{Simulation study 1: Simple correlation structure}
\label{res:study1}

\begin{sidewaysfigure}[!htbp]
	\centering
	\includegraphics[scale=0.6]{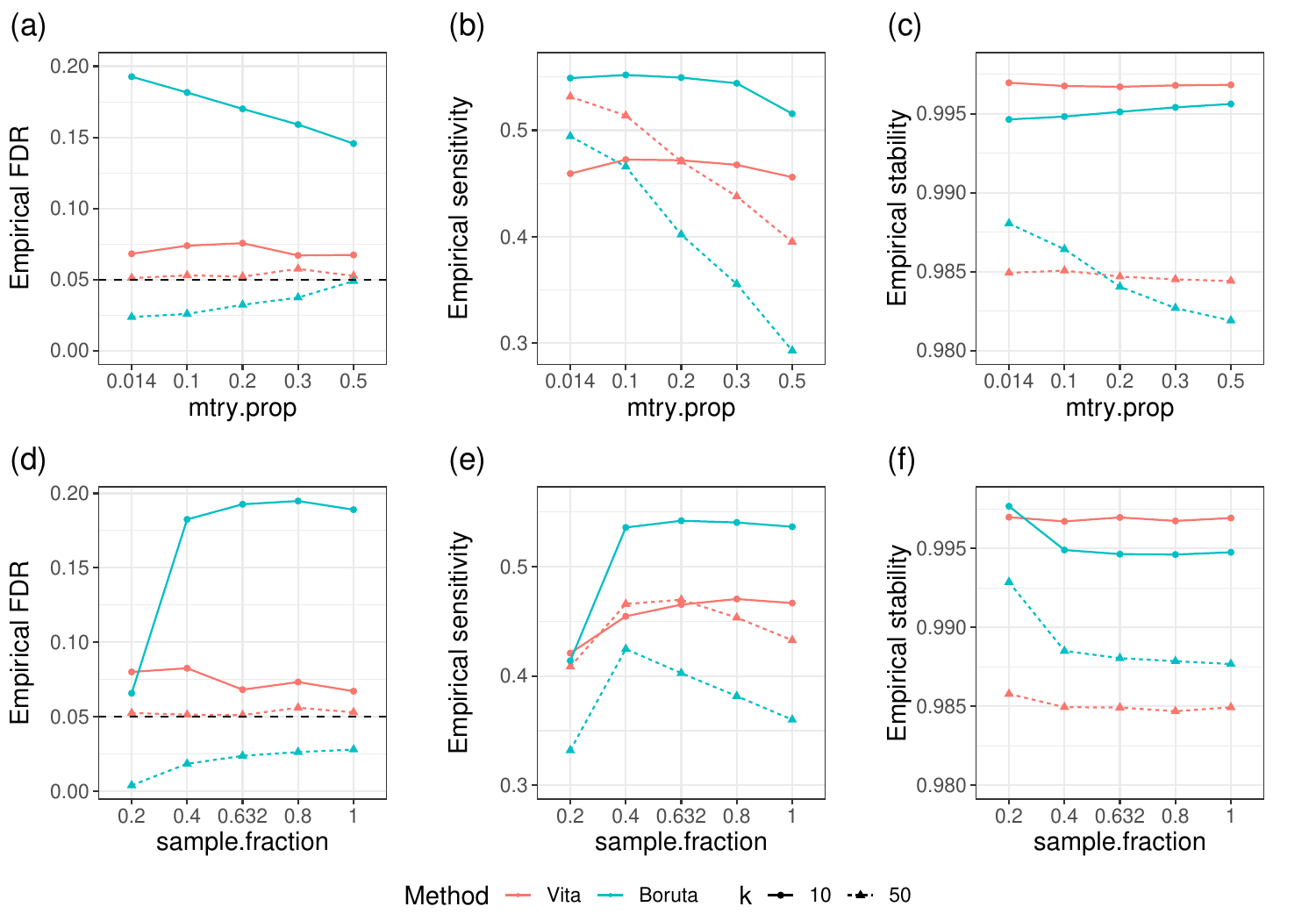}
	\caption{\emph{Simulation study 1: Empirical performances of Vita and Boruta for variations of \texttt{mtry.prop} and \texttt{sample.fraction} are shown in the first and second rows, respectively. For each method, the average across all replicates is shown.}}
	\label{fig:mtrySampFrac01}
\end{sidewaysfigure}

The aim of simulation study 1 is to investigate the influence of the RF hyperparameters on Vita and Boruta, given a simple correlation structure of predictor variables. Figure \ref{fig:mtrySampFrac01} shows the impact of the proportion of split candidates (first row) and sample fraction (second row) on the variable selection procedures.

\hypertarget{sim1:mtry}{
\subsubsection{Simulation study 1: Influence of \texttt{mtry.prop}}\label{sim1:mtry}}

The impact of \texttt{mtry.prop} on the FDR (figure \ref{fig:mtrySampFrac01}-(a)) is different for the two approaches. For Vita, the empirical FDR is similar across different values of \texttt{mtry.prop}, while for Boruta, it decreases with increasing \texttt{mtry.prop} values for $k = 10$, but increases for $k = 50$. Of note, for $k=10$ both methods have an average FDR larger than 0.05; however, it is substantially larger for Boruta and attributable to the stronger correlation structure. The impact of \texttt{mtry.prop} on the sensitivity (figure \ref{fig:mtrySampFrac01}-(b)) also depends on the correlation levels. The sensitivity of both selection procedures decreases more rapidly for the weaker correlated predictor variables ($k = 50$) than for stronger correlated ones ($k = 10$). Apart from the sensitivity of Vita for ($k = 10$), the empirical sensitivity decreases with increasing values of \texttt{mtry.prop}.
The stability of the two selection procedures is very high across all parameter combinations and thus is not strongly affected by \texttt{mtry.prop} (figure \ref{fig:mtrySampFrac01}-(c)).

\hypertarget{sim1:sample.fraction}{
\subsubsection{Simulation study 1: Influence of \texttt{sample.fraction}}\label{sim1:sample.fraction}}

Regarding the empirical FDR (figure \ref{fig:mtrySampFrac01}-(d)), the impact of \texttt{sample.fraction} on Vita is very weak. However, for Boruta, and for $k = 50$, the empirical FDR strictly increases with an increasing \texttt{sample.fraction}. For $k = 10$, we observe a large difference in the FDR when using a value of $0.2$ versus $0.4$. Only for $0.2$ the FDR is close to 0.05.
The sensitivity of both Vita and Boruta (figure \ref{fig:mtrySampFrac01}-(e)) first increases for all correlation levels until some maximum value and then monotonously decreases, where the decrease is more rapid for $k = 50$ than for $k = 10$. It is also important to mention that the maximal sensitivity is not always observed for the default value of \texttt{sample.fraction}.
Again, stability is not strongly influenced by \texttt{sample.fraction} (\ref{fig:mtrySampFrac01}-(f)).

\hypertarget{sim1:min.node.size.prop}{
\subsubsection{Simulation study 1: Influence of \texttt{min.node.size.prop} and \texttt{replace}}\label{sim1:min.repl}}

Concerning the impact of the minimal terminal node size proportion and the resampling strategy, the observed influences are negligible (figure \ref{fig:minReplace01}).

\subsection{Simulation study 2: Empirical correlation structure}
\label{res:study2}
The aim of this study is analogous to that of the previous study, which is to investigate the influences of the hyperparameters on the Vita and the Boruta variable selection methods. However, the data in this simulation study differ from those in study 1 since the predictor variables are based on experimental gene expression data, and the noise variables are uncorrelated.

\hypertarget{sim2:mtryRes}{
\subsubsection{Simulation study 2: Influence of mtry.prop}\label{mtryRes2}}

\begin{figure}[!htbp]
	\centering
	\includegraphics[scale=0.6]{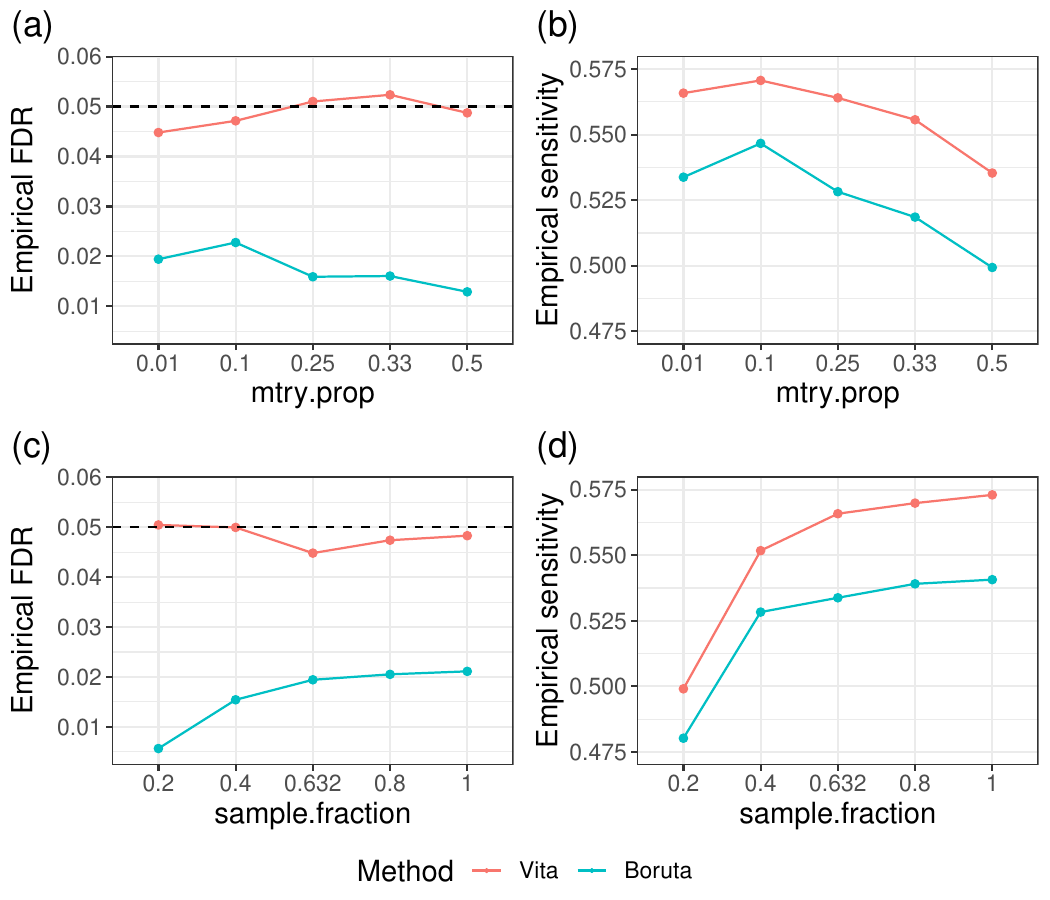}
	\caption{Simulation study 2: Empirical performances of Vita and Boruta for variations of \texttt{mtry.prop} and \texttt{sample.fraction} are shown in the first and second rows, respectively. Each method's average across all replicates is shown for each hyperparameter variation.}
	\label{fig:mtrySampFrac02}
\end{figure}

The influence of \texttt{mtry.prop} on the empirical FDR of both approaches is small (figure \ref{fig:mtrySampFrac02}-(a)) with FDR for Vita varying around $0.05$ and for Boruta around $0.02$. Vita and Boruta's empirical sensitivity first increases and then strictly decreases (figure \ref{fig:mtrySampFrac02}-(b)). The maximum sensitivity of both selection procedures is obtained for the same \texttt{mtry.prop} value of $0.1$. Overall, the differences in the evaluation criteria between different values of \texttt{mtry.prop} are much smaller than in simulation study 1.  

\hypertarget{sim2:samplefracRes}{
\subsubsection{Simulation study 2: Influence of sample.fraction}\label{sampleFracRes2}}

The effect of \texttt{sample.fraction} on the empirical FDR is similar to the results of $k=50$ in simulation study 1. With increasing values of \texttt{sample.fraction}, the FDR of Boruta increases, but for Vita, the influence of \texttt{sample.fraction} is negligible (\ref{fig:mtrySampFrac02}-(c)). The empirical sensitivity of the two methods is monotonous, increasing as the sample fraction used for training increases (\ref{fig:mtrySampFrac02}-(d)). Again, the differences are smaller than in simulation study 1.  

\hypertarget{sim2:min.node-replace}{
\subsubsection{Simulation study 2: Influence of \texttt{min.node.size.prop} and \texttt{replace}}\label{sim2:min.repl}}

As in simulation study 1, the impact of the minimal terminal node size proportion and the resampling strategy on the observed influences are negligible (figure \ref{fig:minReplace02}).

\subsection{Summary of results for studies 1 and 2}

The results of studies 1 and 2 are summarized in table \ref{chap:hyperparam:sim1and2}. The first remark is that, in many cases, depending on the performance measure, the suitable hyperparameter settings for variable selection differ from the default values for both Vita and Boruta. The difference between default and suitable settings is related to the data structure. Secondly, the correlation structure considerably affects the optimal hyperparameter settings of \texttt{mtry.prop} and \texttt{sample.fraction}. For example, for better sensitivity, small values of the hyperparameters \texttt{mtry.prop} and \texttt{sample.fraction} are preferable for weakly correlated predictor variables for both variable selection methods. Drawing the training dataset with replacement is, in most cases, appropriate for controlling the FDR. Finally, although the minimal terminal node size slightly influences the variable selection procedures, its adequate setting in study 2 differs from the default value.

\begin{table}[htbp!]
	\centering
	\resizebox{\columnwidth}{!}{ 
  \begin{threeparttable}
	\caption{Summary results for studies 1 and 2. The table reports, for numeric hyperparameters, differences between the default hyperparameter settings and those that yield optimal performance measures. For example, the column Difference (Sensitivity) indicates the deviation of the optimal numeric hyperparameter's setting for the corresponding performance measure from the default setting. Entries in bold indicate that the optimal setting is the default value. Entries of the hyperparameter \texttt{replace} correspond to its optimal settings.}
	\begin{tabular}{lllllr}
		\toprule
		\multirow{2}{*}{\textbf{Study}} & \multirow{2}{*}{\textbf{Hyperparam.}} & \multicolumn{2}{c}{\textbf{Difference (Sensitivity)}} & \multicolumn{2}{c}{\textbf{Difference (FDR)}} \\
		\cline{3-4}\cline{5-6}
			 		   & 						 & \textbf{Vita} & \textbf{Boruta} & \textbf{Vita} & \textbf{Boruta} \\
		\toprule
								& \texttt{mtry.prop} 			& $0.086\ (0.47)$ & $0.086\ (0.55)$ 												& $0.286\ (0.07)$ & $0.486\ (0.14)$ \\
				1 ($k = 10$)  	& \texttt{sample.fraction}  	& $0.168\ (0.47)$ & $\textbf{0.00\ (0.54)}$ 								& $0.368\ (0.07)$ & $0.432\ (0.07)$ \\
								& \texttt{replace}  			& $\textbf{\texttt{TRUE}}\ \mathbf{(0.46)}$ & $\textbf{\texttt{TRUE}}\ \mathbf{(0.54)}$						& $\textbf{\texttt{TRUE}}\ \mathbf{(0.07)}$ & $\textbf{\texttt{TRUE}}\mathbf{(0.2)}$\\
								& \texttt{min.node.size.prop} 	& $\mathbf{0.00\ (0.46)}$ & $0.04\ (0.55)$ 									& $\mathbf{0.00\ (0.07)}$ & $\mathbf{0.00\ (0.19)}$ \\
		\midrule
								& \texttt{mtry.prop} 			& $\mathbf{0.00\ (0.53)}$ & $\textbf{0.00\ (0.49)}$ 						& $\textbf{0.00\ (0.05)}$ & $\textbf{0.00\ (0.02)}$ \\
		 		1 ($k = 50$)  	& \texttt{sample.fraction} 	& $\mathbf{0.00\ (0.47)}$ & $0.232\ (0.42)$ 									& $0.432\ (0.06)$ & $0.432\ (0.004)$ \\
								& \texttt{replace}  			& $\textbf{\texttt{TRUE}\ (0.47)}$ & $\texttt{FALSE}\ (0.4)$ 				& $\textbf{\texttt{TRUE}\ (0.05)}$ & $\textbf{\texttt{TRUE}}\ \mathbf{(0.02)}$\\
								& \texttt{min.node.size.prop} 	& $\mathbf{0.00\ (0.53)}$ & $0.09\ (0.50)$ 									& $\mathbf{0.00\ (0.05)}$ & $\mathbf{0.00\ (0.02)}$ \\
		\midrule
								& \texttt{mtry.prop} 			& $0.09\ (0.57)$ & $0.09\ (0.54)$ 												& $\mathbf{0.00\ (0.04)}$ & $\textbf{0.00\ (0.01)}$ \\
				2 				& \texttt{sample.fraction} 		& $.368\ (0.57)$ & $.368\ (0.54)$												& $\mathbf{0.00\ (0.04)}$ & $0.432\ (0.005)$ \\
								& \texttt{replace}  			& $\textbf{\texttt{TRUE}\ (0.56)}$ & $\textbf{\texttt{TRUE}\ (0.53)}$	& $\textbf{\texttt{TRUE}\ (0.05)}$ & $\textbf{\texttt{TRUE}\ (0.2)}$ \\
								& \texttt{min.node.size.prop} 	& $0.05\ (0.57)$ & $0.19\ (0.53)$ 												& $0.19\ (0.05)$ & $0.09\ (0.02)$ \\
		\bottomrule
	\end{tabular}
	\label{chap:hyperparam:sim1and2}

\end{threeparttable}
}
	\label{results:summary}
\end{table}

\FloatBarrier 

\section{Discussion}
\label{sec:disc}

We investigated the impact of the RF hyperparameters on the Vita and the Boruta variable selection methods. In general, the hyperparameters \texttt{mtry.prop} and \texttt{sample.fraction} have a stronger influence on the two variable selection methods than \texttt{replace} and \texttt{min.node.size.prop} and the optimal values depend on the data structure. FDR is not strongly influenced by different values of both hyperparameters, but the default value of \texttt{mtry.prop} is only optimal for weakly correlated predictor variables in simulation study 1. In the other settings, slightly larger values result in better sensitivity. In contrast, the difference in sensitivity of the optimal compared to the default value of \texttt{sample.fraction} is negligible for strongly correlated predictor variables, whereas smaller values than the default are better in the other settings. Since we did not observe any effect of the drawing strategy, we recommend sampling with replacement as in \cite{martinez2010}. A good compromise between the computational time and a minimal node size should be found.

Some of our findings corroborate those previously established regarding variable importance. Similarly to the results of \cite{lunettascreening2004,genuervariable2010} we found that the data structure is essential. For example, the impact of \texttt{mtry.prop} is more important for weakly correlated predictor variables than for strongly correlated ones. The decrease in sensitivity for increasing values of \texttt{mtry.prop} can be explained by the fact that a small number of split candidates favors variables with moderate or smaller effects on the response variable \cite{probsthyperparameters2019,grompingvariable_2009} leading to an increase in their expected importance, as previously shown by \cite{genuervariable2010}. According to the argumentation of \cite{probsthyperparameters2019}, the hyperparameter \texttt{sample.fraction} should be kept low to achieve a better prediction performance. The authors do not investigate the impact of the correlation structure of the setting of this hyperparameter. Our results show that a low \texttt{sample.fraction} value is not always suitable for the variable selection methods and depends on the correlation of predictor variables. Concerning the sampling strategy of the training dataset (\texttt{replace}), the results of \cite{martinez2010} imply that the sampling strategy does not have an impact on the prediction performance for an optimal sample fraction. Similarly, we did not observe an impact of the sampling strategy on the variable selection method. 

We conclude that the default settings of hyperparameters can not necessarily be recommended for variable selection. However, it is difficult to make general recommendations based on our results. It would be necessary to perform additional simulation studies to better understand the influence of \texttt{mtry.prop} and \texttt{sample.fraction} on the sensitivity of variable selection by systematically varying additional characteristics of the datasets such as number of predictor variables and effect sizes. Here we focused on the effect of different correlation structures. Another limitation is that we set all other hyperparameters to their default values apart from the one being analyzed. Some hyperparameters may interact, which could not be analyzed with our design. For example, the sample fraction and drawing strategy are directly related when defining the training dataset for each decision tree. Another example concerns the minimal terminal node size proportion and the proportion of splitting candidate variables. For a given value of \texttt{mtry.prop}, decreasing the minimal terminal node size proportion increases the changes of predictor variables with lower effects on the response variable for splitting. Finally, we simulated classification problems only and regression settings have not been considered. The reason is that no study has so far systematically assessed a corrected importance for quantitative response variables. Using this performance measure without a prior validation study could result in spurious findings.

One important open question remains after this study. It is currently unclear how to select the optimal hyperparameter values for analyzing a specific dataset. Usually, hyperparameters are tuned in the context of developing a prediction model with optimal prediction performance. Thus, this criterion can be used to evaluate different parameter combinations within a resampling scheme such as cross-validation. However, it will not be useful if the goal is to select all relevant predictor variables since different sets of predictor variables often lead to very similar performance \cite{degenhardtevaluation2019}. Two of the evaluation criteria used in this study, FDR and sensitivity, could only be estimated since we analyzed simulated data where the truth is known. It would be possible to calculate stability on experimental datasets using for example a bootstrapping approach, but we did not observe any relevant differences across different hyperparameter combinations. 

\backmatter

\bmhead{Code and data availibility}

\texttt{R}-Code to reproduce results presented in this article is available in \url{https://github.com/imbs-hl/RF-hyperparameters-and-variable-selection}. Simulated data are available in \url{https://zenodo.org/record/8308235} under DOI: 10.5281/zenodo.8308235.
\bmhead{Acknowledgments}

This work was funded by the German Center for Cardiovascular Research (Bundesministerium für Bildung und Forschung, grant 81Z1700103) and the German Research Foundation (DFG, grant \# KO2240).

\newpage

\FloatBarrier
\newpage

\bibliography{bibliography}

\begin{thebibliography}{27}
\providecommand{\natexlab}[1]{#1}
\providecommand{\url}[1]{{#1}}
\providecommand{\urlprefix}{URL }
\providecommand{\doi}[1]{\url{https://doi.org/#1}}
\providecommand{\eprint}[2][]{\url{#2}}
 \bibcommenthead

\bibitem[{Benjamini and Hochberg(1995)}]{benjaminicontrolling1995}
Benjamini Y, Hochberg Y (1995) Controlling the {False} {Discovery} {Rate}: {A}
  {Practical} and {Powerful} {Approach} to {Multiple} {Testing}. Journal of the
  Royal Statistical Society: Series B (Methodological) 57(1):289--300.
  \doi{10.1111/j.2517-6161.1995.tb02031.x},
  \urlprefix\url{https://onlinelibrary.wiley.com/doi/10.1111/j.2517-6161.1995.tb02031.x}

\bibitem[{Bland and Altman(1995)}]{blandmultiple1995}
Bland JM, Altman DG (1995) Multiple significance tests: the {Bonferroni}
  method. BMJ 310(6973):170. \doi{10.1136/bmj.310.6973.170},
  \urlprefix\url{https://www.bmj.com/content/310/6973/170}

\bibitem[{Boulesteix et~al(2012)Boulesteix, Bender, Lorenzo~Bermejo, and
  Strobl}]{boulesteixrandom2012}
Boulesteix AL, Bender A, Lorenzo~Bermejo J, et~al (2012) Random forest {Gini}
  importance favours {SNPs} with large minor allele frequency: impact, sources
  and recommendations. Briefings in Bioinformatics 13(3):292--304.
  \doi{10.1093/bib/bbr053}

\bibitem[{Breiman(2001)}]{breimanrandom2001}
Breiman L (2001) Random {Forests}. Machine Learning 45(1):5--32.
  \doi{10.1023/A:1010933404324},
  \urlprefix\url{https://doi.org/10.1023/A:1010933404324}

\bibitem[{Budczies and Kosztyla(2020)}]{cancerdata}
Budczies J, Kosztyla D (2020) cancerdata: Development and validation of
  diagnostic tests from high-dimensional molecular data: Datasets. R package
  version 1.28.0

\bibitem[{Degenhardt et~al(2019)Degenhardt, Seifert, and
  Szymczak}]{degenhardtevaluation2019}
Degenhardt F, Seifert S, Szymczak S (2019) Evaluation of variable selection
  methods for random forests and omics data sets. Briefings in Bioinformatics
  20(2):492--503. \doi{10.1093/bib/bbx124}

\bibitem[{D\'iaz-Uriarte and Alvarez~de Andr\'es(2006)}]{diazuriartegene2006}
D\'iaz-Uriarte R, Alvarez~de Andr\'es S (2006) Gene selection and
  classification of microarray data using random forest. BMC Bioinformatics
  7(1):3. \doi{10.1186/1471-2105-7-3},
  \urlprefix\url{https://doi.org/10.1186/1471-2105-7-3}

\bibitem[{Genuer et~al(2010)Genuer, Poggi, and
  Tuleau-Malot}]{genuervariable2010}
Genuer R, Poggi JM, Tuleau-Malot C (2010) Variable selection using random
  forests. Pattern Recognition Letters 31(14):2225--2236.
  \doi{10.1016/j.patrec.2010.03.014},
  \urlprefix\url{https://www.sciencedirect.com/science/article/pii/S0167865510000954}

\bibitem[{Gr\"omping(2009)}]{grompingvariable_2009}
Gr\"omping U (2009) Variable {Importance} {Assessment} in {Regression}:
  {Linear} {Regression} versus {Random} {Forest}. The American Statistician
  63(4):308--319. \doi{10.1198/tast.2009.08199},
  \urlprefix\url{https://doi.org/10.1198/tast.2009.08199}

\bibitem[{He and Yu(2010)}]{hestable2010}
He Z, Yu W (2010) Stable feature selection for biomarker discovery.
  Computational Biology and Chemistry 34(4):215--225.
  \doi{10.1016/j.compbiolchem.2010.07.002},
  \urlprefix\url{https://www.sciencedirect.com/science/article/pii/S1476927110000502}

\bibitem[{Ishwaran et~al(2008)Ishwaran, Kogalur, Blackstone, and
  Lauer}]{ishwaranRSF2008}
Ishwaran H, Kogalur UB, Blackstone EH, et~al (2008) Random survival forests.
  The Annals of Applied Statistics 2(3):841--860. \doi{10.1214/08-AOAS169},
  \urlprefix\url{https://projecteuclid.org/journals/annals-of-applied-statistics/volume-2/issue-3/Random-survival-forests/10.1214/08-AOAS169.full}

\bibitem[{Janitza et~al(2018)Janitza, Celik, and Boulesteix}]{janitza2016}
Janitza S, Celik E, Boulesteix AL (2018) A computationally fast variable
  importance test for random forests for high-dimensional data. Advances in
  Data Analysis and Classification 12(4):885--915.
  \doi{10.1007/s11634-016-0276-4},
  \urlprefix\url{https://doi.org/10.1007/s11634-016-0276-4}

\bibitem[{Kursa and Rudnicki(2010)}]{kursafeature2010}
Kursa MB, Rudnicki WR (2010) Feature {Selection} with the {Boruta} {Package}.
  Journal of Statistical Software 36:1--13. \doi{10.18637/jss.v036.i11},
  \urlprefix\url{https://doi.org/10.18637/jss.v036.i11}

\bibitem[{Lang et~al(2017)Lang, Bischl, and Surmann}]{langbatchtools2017}
Lang M, Bischl B, Surmann D (2017) batchtools: {Tools} for {R} to work on batch
  systems. Journal of Open Source Software 2(10):135.
  \doi{10.21105/joss.00135},
  \urlprefix\url{https://joss.theoj.org/papers/10.21105/joss.00135}

\bibitem[{Liaw and Wiener(2002)}]{randomforest}
Liaw A, Wiener M (2002) Classification and regression by randomforest. R News
  2(3):18--22. \urlprefix\url{https://CRAN.R-project.org/doc/Rnews/}

\bibitem[{Lunetta et~al(2004)Lunetta, Hayward, Segal, and
  Van~Eerdewegh}]{lunettascreening2004}
Lunetta KL, Hayward LB, Segal J, et~al (2004) Screening large-scale association
  study data: exploiting interactions using random forests. BMC genetics 5:32.
  \doi{10.1186/1471-2156-5-32}

\bibitem[{Michiels et~al(2005)Michiels, Koscielny, and Hill}]{michiels2005}
Michiels S, Koscielny S, Hill C (2005) Prediction of cancer outcome with
  microarrays: a multiple random validation strategy. Lancet (London, England)
  365(9458):488--492. \doi{10.1016/S0140-6736(05)17866-0}

\bibitem[{Morris et~al(2019)Morris, White, and Crowther}]{morris2019}
Morris TP, White IR, Crowther MJ (2019) Using simulation studies to evaluate
  statistical methods. Statistics in Medicine 38(11):2074--2102.
  \doi{10.1002/sim.8086},
  \urlprefix\url{https://onlinelibrary.wiley.com/doi/10.1002/sim.8086}

\bibitem[{Mart\'{i}\~{n}ez Mu\~{n}oz and Su\'arez(2010)}]{martinez2010}
Mart\'{i}\~{n}ez Mu\~{n}oz G, Su\'arez A (2010) Out-of-bag estimation of the
  optimal sample size in bagging. Pattern Recognition 43(1):143--152.
  \doi{10.1016/j.patcog.2009.05.010},
  \urlprefix\url{https://www.sciencedirect.com/science/article/pii/S003132030900212X}

\bibitem[{Nembrini et~al(2018)Nembrini, König, and
  Wright}]{nembrinirevival2018}
Nembrini S, König IR, Wright MN (2018) The revival of the {Gini} importance?
  Bioinformatics 34(21):3711--3718. \doi{10.1093/bioinformatics/bty373}

\bibitem[{Oshiro et~al(2012)Oshiro, Perez, and Baranauskas}]{oshiro2012}
Oshiro TM, Perez PS, Baranauskas JA (2012) How {Many} {Trees} in a {Random}
  {Forest}? In: Perner P (ed) Machine {Learning} and {Data} {Mining} in
  {Pattern} {Recognition}. Springer, Berlin, Heidelberg, Lecture {Notes} in
  {Computer} {Science}, pp 154--168, \doi{10.1007/978-3-642-31537-4_13}

\bibitem[{Probst et~al(2019)Probst, Wright, and
  Boulesteix}]{probsthyperparameters2019}
Probst P, Wright MN, Boulesteix AL (2019) Hyperparameters and tuning strategies
  for random forest. WIREs Data Mining and Knowledge Discovery 9(3):e1301.
  \doi{10.1002/widm.1301},
  \urlprefix\url{https://onlinelibrary.wiley.com/doi/abs/10.1002/widm.1301}

\bibitem[{{Scornet, Erwan}(2017)}]{refId0}
{Scornet, Erwan} (2017) Tuning parameters in random forests. ESAIM: Procs
  60:144--162. \doi{10.1051/proc/201760144},
  \urlprefix\url{https://doi.org/10.1051/proc/201760144}

\bibitem[{Segal(2004)}]{segal2004}
Segal MR (2004) Machine {Learning} {Benchmarks} and {Random} {Forest}
  {Regression} \urlprefix\url{https://escholarship.org/uc/item/35x3v9t4}

\bibitem[{Strobl et~al(2007)Strobl, Boulesteix, Zeileis, and
  Hothorn}]{stroblbias2007}
Strobl C, Boulesteix AL, Zeileis A, et~al (2007) Bias in random forest variable
  importance measures: {Illustrations}, sources and a solution. BMC
  Bioinformatics 8(1):25. \doi{10.1186/1471-2105-8-25},
  \urlprefix\url{https://doi.org/10.1186/1471-2105-8-25}

\bibitem[{van~'t Veer et~al(2002)van~'t Veer, Dai, van~de Vijver, He, Hart,
  Mao, Peterse, van~der Kooy, Marton, Witteveen, Schreiber, Kerkhoven, Roberts,
  Linsley, Bernards, and Friend}]{veer2002}
van~'t Veer LJ, Dai H, van~de Vijver MJ, et~al (2002) Gene expression profiling
  predicts clinical outcome of breast cancer. Nature 415(6871):530--536.
  \doi{10.1038/415530a},
  \urlprefix\url{https://www.nature.com/articles/415530a}

\bibitem[{{Wright} and Ziegler(2017)}]{ranger}
{Wright} MN, Ziegler A (2017) {ranger}: A fast implementation of random forests
  for high dimensional data in {C++} and {R}. Journal of Statistical Software
  77(1):1--17. \doi{10.18637/jss.v077.i01}

\end{thebibliography}

\begin{appendices}
\begin{sidewaysfigure}[!htbp]
\section{Supplementary figures}\label{appendix}

Supplementary figures of simulation study 1 and simulation study 2 are presented in this section.

\subsection{Simulation study 1}
\label{sim1appendix}

	\centering
	\includegraphics[scale=0.6]{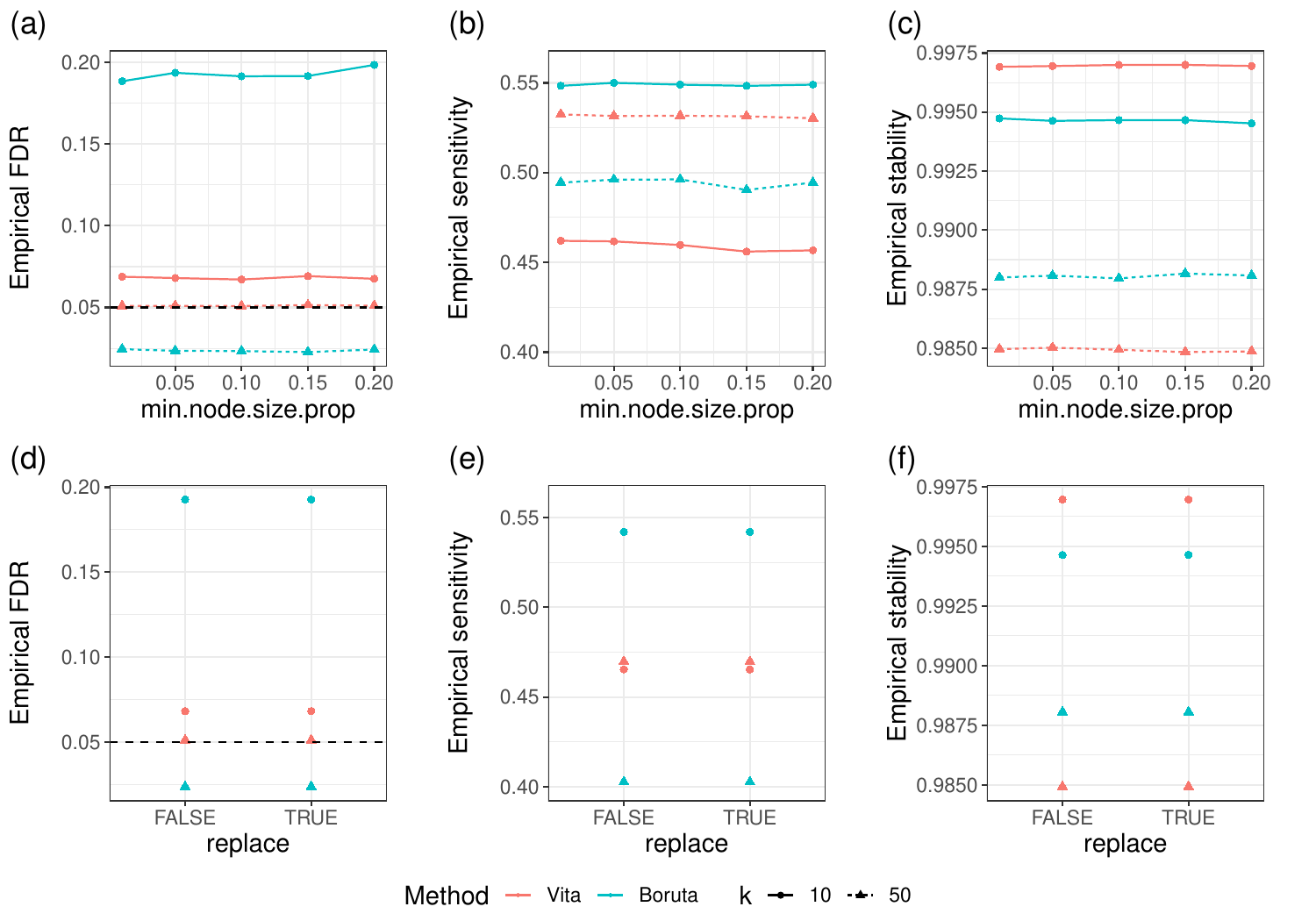}
		\caption{\emph{Simulation study 1: the first row shows the empirical performances of Vita and Boruta for variations of \texttt{min.node.size.prop}, and their second row the empirical performances for variations of \texttt{replace}. Axes are scaled differently.}} 
	\label{fig:minReplace01}
\end{sidewaysfigure}

\newpage
\FloatBarrier
\subsection{Simulation study 2}
\begin{figure}[!htbp]
	\centering
	\includegraphics[scale=0.6]{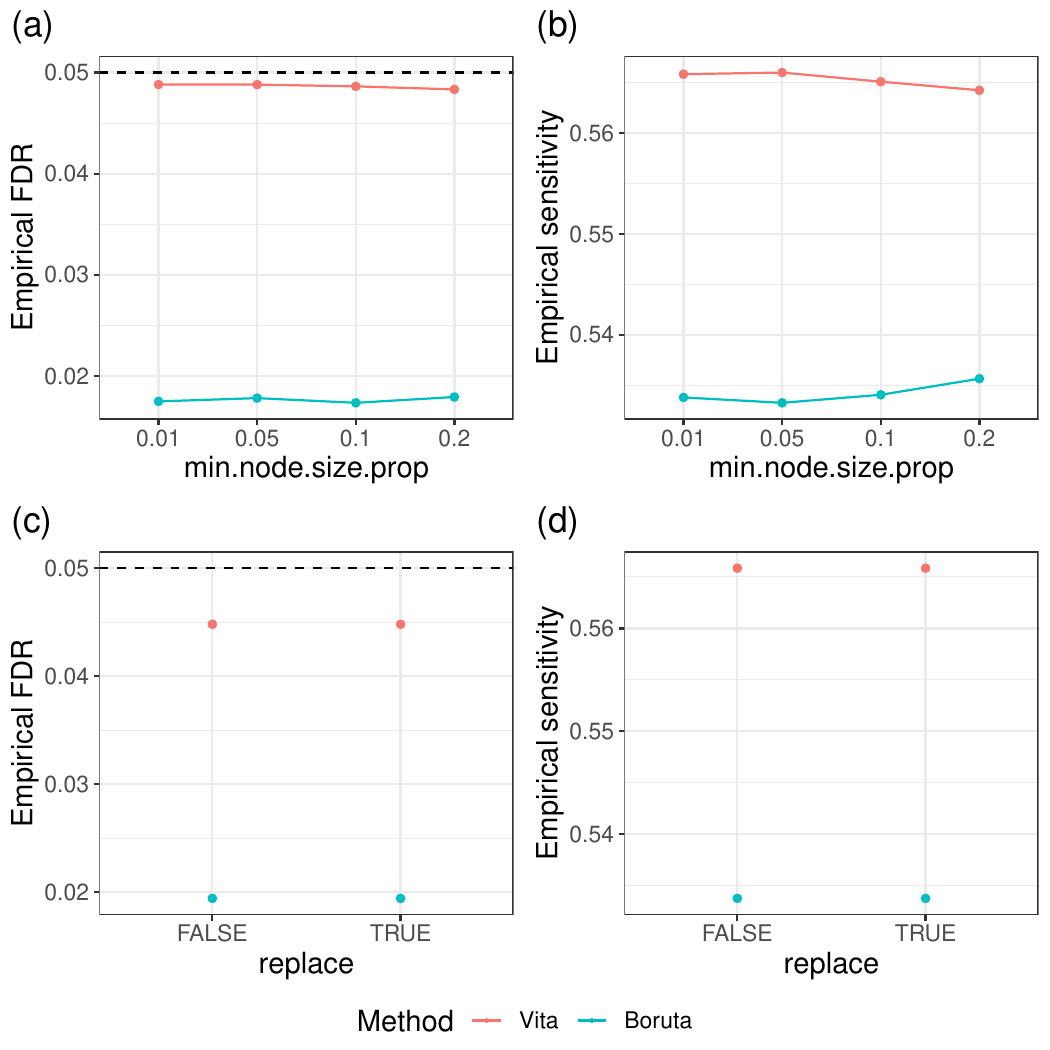}
	\caption{\emph{Simulation study 2: the first row shows the empirical performances of Vita and Boruta for variations of \texttt{min.node.size.prop}, and their second row the empirical performances for variations of \texttt{replace}. Axes are scaled differently.}}
	\label{fig:minReplace02}
\end{figure}

\end{appendices}

\end{document}